\documentclass{article}
\usepackage{acl}
\usepackage[T1]{fontenc} 
\usepackage{times}
\usepackage{inconsolata}
\usepackage{xcolor}
\usepackage{hyperref}
\usepackage{amsmath}
\usepackage{tikz}
\usepackage{float}
\usepackage{booktabs}
\usepackage{placeins}
\usepackage{listings}
\usepackage{microtype}
\usepackage{stfloats}\fnbelowfloat 
\newcommand{\nAgents}{five}
\newcommand{\nPersonas}{eight}
\newcommand{\nPerPersona}{five}
\newcommand{\nModels}{ten}

\title{Back in Style: A Sociolinguistic Approach to Authoring and Measuring Persona Fidelity in User Simulation \\
\texorpdfstring{%
  \raisebox{-0.2\height}{\includegraphics[height=7mm]{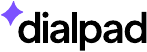}}%
}{}}
\author{Lex Konnelly, Elena Khasanova, Riqiang Wang, \\
\textbf{Matthias Lee, Harsh Saini, Parsa Kavehzadeh} \\ \texttt{\{lex.konnelly, elena.khasanova, riqiang.wang\}}\texttt{@dialpad.com}, \\ \texttt{\{matthias.lee, parsa.kavehzadeh}\}\texttt{@dialpad.com}\\{Dialpad Inc.}}
\date{August 2026}

\begin{document}

\maketitle

\begin{abstract}
As agentic systems gain commercial popularity, user simulators increasingly serve as measurement instrument for their evaluation. However, the fidelity of simulated users in comparison to real human users is generally low, and typically assessed by costly, subjective LLM judges. In this pilot study, we ask whether fidelity can instead be measured deterministically by treating a user persona \emph{sociolinguistically}: as a social type that emerges from observable linguistic style, rather than one predicted by labels or descriptions a model must extrapolate into behaviour. We author personas as concrete stylistic rates, which lets us transfer two established, model-free instruments --- authorship-verification stylometry and lexicon-based content analysis --- as fidelity diagnostics. We A/B-test the sociolinguistic schema against a flat descriptive baseline across \nAgents{} task-oriented customer-service agents. Results show that the sociolinguistic schema improves both stylistic adherence and stylometric distinguishability for most of the tested models, with a caveat that persona style fidelity does not necessarily equal persona ``naturalness''. We argue that a sociolinguistic approach to persona design is a promising path towards more diverse and representative user personas, and that these metrics are most valuable in an error-attribution analysis, localizing where fidelity breaks down. This is a first step towards interventions that move user simulations closer to faithful renderings of diverse and variable linguistic outputs.
\end{abstract}

\section{Introduction}

User simulators are becoming a standard measurement instrument for evaluating agentic systems  \citep{abdulhai2025consistently, liu2026muse, hathidara2026mirrorbench}. For production evaluation systems, a simulated user exercises the full system end to end at scale: it drives complete multi-turn interactions through the agent's tool calls and the downstream side effects they produce, which can surface integration and data-flow errors that single-turn or unit tests alone do not reach. However, simulation also introduces its own conversational artifacts and failure patterns unlike those of real users, and evaluating agents on simulated users can misrepresent real-world capability, especially across diverse user populations  \citep{seshadri2026lost, zhou2026sim2real}. This makes the simulator's own fidelity a core priority in agentic evaluation pipelines.  

This is further complicated by how personas are conventionally specified (i.e. what instructions are supplied to the simulator for its rendering of a given user), typically as lists of adjectives or demographic labels the simulator must extrapolate into behaviour. We argue that a sociolinguistic approach to \textit{persona} schema design offers a promising path to designing more faithful and diverse user personas. Where labels or prose-level descriptions treat identity as \emph{predictive of} behaviour (the model infers how a specified type, e.g. "a confused client", would act), a sociolinguistic approach treats identity as \emph{emergent from} linguistic expression. In this view, a \textit{persona} is a social type constituted by recurrent stylistic choices and discursive stance-taking \citep{dubois2007stance, kiesling2009style, kiesling2018interactional, ochs1993constructing}. Treating observable linguistic style as the substrate lets us author a user persona as concrete, measurable features such as rates of hedging, politeness, filler words, and emoji or internet-speak usage. This also makes persona fidelity legible to model-free, spec-mapped metrics.

Testing this grounding poses two coupled questions: (a) does persona specification as linguistic style measurably improve a user simulator's fidelity, and (b) can that improvement be read off deterministically, without requiring a calibrated LLM judge or human annotator? We treat the second as fundamentally a problem of metric transfer, taking two established deterministic instruments (authorship-verification stylometry \citep{aggazzotti2023speech, huang2024authorship} and lexicon-based content analysis \citep{hutto2014vader, tausczik2010liwc}) and applying them on the style substrate as an exploratory fidelity diagnostic. We test whether they yield actionable signal: can they localize which components of a persona's linguistic structure a simulator does and does not render faithfully? Such localization can then be used as a lever: targeted behavioural interventions on simulated users (e.g. instructing them to limit politeness) have been shown to mitigate gaps for the most miscalibrated user groups \citep{seshadri2026lost}. 

We report results from a a pilot experiment applying both metrics to a style-based \emph{definition} schema against a flat label-based \emph{archetype} baseline across \nAgents{} task-oriented customer-service agents in text and voice modes, with {\nModels} candidate models (\texttt{gpt-4o-mini}, \texttt{gpt-5-mini}, \texttt{gpt-4o-realtime-preview}, \texttt{gemini-3.5-flash (low)} \footnote{"low"/"high" refer to Gemini's thinking\_level setting.}, \texttt{gemini-3.5-flash (high)}, \texttt{gemini-3.1-pro-preview}, \texttt{gemma-4-31b-it}, \texttt{claude-opus-5}, \texttt{claude-sonnet-5}, and \texttt{claude-haiku-4-5}) behind the user simulator. These models were chosen because they represent a mixture of providers, capabilities, and open and closed source alternatives. The scores indicate that the sociolinguistically-grounded persona design improved faithfulness of a persona's linguistic style in comparison to the baseline descriptive prompt for a set of specified stylistic features, which resulted in a greater inter-persona distinguishability for the majority of models under test. These effects are strongest for \texttt{gpt-5-mini} and also hold under the voice modality; \texttt{\texttt{gemini-3.1-pro-preview} and gemini-3.5-flash}, regardless of reasoning effort, showed little effect between schemas, suggesting both that a sufficiently steerable simulator is needed to get a strong effect, and a stronger model $\neq$ better results; instead, model family with its specific preference alignment is more predictive of schema effect than relative capability. Qualitative review of the data also showed that fidelity to the persona schema $\neq$ naturalness, and further work is needed to dial in more natural-sounding stylistic variation for user personas. We provide an overview of proposed future directions, and highlight the promising aspects of our approach for narrowing gaps in faithful rendering of more diverse human language input.  

Core contributions include: 
\begin{enumerate}
  \item A sociolinguistically-motivated persona schema treating persona as accreted stance, realized as style, specified as modality-specific stylistic \emph{rates} as an alternative to adjective lists or demographic labels.
  \item Two deterministic, spec-mapped fidelity diagnostics adapted from prior art and applied as primary fidelity signals in a task-oriented customer-service simulator across text and voice: (a)~stylometric distinguishability (authorship-verification ROC-AUC), and (b)~stylistic adherence (spec-ordered vs.\ observed trait magnitudes, Kendall's $\tau_b$).
  \item A failure-source decomposition under metric transfer: holding the metrics fixed as an instrument, we separate two potential sources of low fidelity: spec/schema underspecification and a model-steerability ceiling, via a schema$\times$model factorial in text, and test whether the schema effect replicates under a voice pipeline.
\end{enumerate}

\section{Related work}
\label{sec:related}

\subsection{Defining a user persona for simulation}
\label{sec:related-persona}

Most task-oriented LLM-based user simulators converge on one of two types of persona steering. The first is to specify a persona as a profile or attribute list, e.g. goals, demographics, or personality tags, which the model is asked to render into user behaviour \citep{chen2025recusersim, liu2026muse, du2025twinvoice, abdulhai2025consistently}. The second representation steers a single trait along a control knob and checks that the output moves with it \citep{chen2025recusersim, han2026personality}; role-steering studies show that LLM personas can be pushed along social dimensions \citep{song2026rolesteering}, but also collapse back toward a default assistant voice under sustained interaction \citep{xiao2026chameleon, lu2026assistantaxis}. Both representations share a gap, as they do not specify explicitly the quantity that a fidelity metric must measure.

Sociolinguistic research has long treated surface-level linguistic forms as exactly this kind of observable quantity. In sociolinguistic terms, a \textit{persona} is not simply a collection of speaker qualities, but a social type that emerges from recurrent choices of linguistic expression. A speaker takes \textit{stances} in discourse – evaluative, affective, and epistemic positions toward the topic of talk and their interlocutor – and as these stances recur, they accrete into a durable linguistic style that constitutes the persona  \citep{dubois2007stance, kiesling2009style, kiesling2018interactional, ochs1993constructing}. Under this account, style is the medium in which the persona itself is constituted: the social type resides in observable language, rather than behind it. 

\begin{figure}[ht]
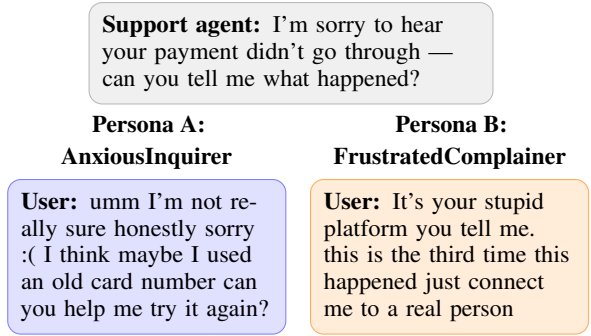

\centering
\newcommand{\chatbubble}[3]{
  \tikz\node[rounded corners=5pt, inner sep=5pt, text width=.90\linewidth,
             align=left, fill=#1, draw=#2, line width=.4pt, font=\footnotesize]{#3};}
\begin{minipage}{.72\linewidth}
\chatbubble{gray!12}{gray!55}{\textbf{Support agent:} I'm sorry to hear your payment didn't go through --- can you tell me what happened?}
\end{minipage}

\smallskip
\begin{minipage}[t]{.48\linewidth}
\centering
{\footnotesize\textbf{Persona A: AnxiousInquirer}}\par\smallskip
\chatbubble{blue!12}{blue!50}{\textbf{User:} umm I'm not really sure honestly sorry :( I think maybe I used an old card number can you help me try it again?}
\end{minipage}\hfill
\begin{minipage}[t]{.48\linewidth}
\centering
{\footnotesize\textbf{Persona B: FrustratedComplainer}}\par\smallskip
\chatbubble{orange!15}{orange!65}{\textbf{User:} It's your stupid platform you tell me. this is the third time this happened just connect me to a real person}
\end{minipage}
\caption{\small From an identical prompt, the \emph{AnxiousInquirer} and \emph{FrustratedComplainer} personas diverge into opposite styles. \citet{dubois2007stance}'s \emph{stance triangle} holds that an utterance at once performs three actions: \emph{evaluates} the object of talk, \emph{positions} the speaker's affect and certainty, and \emph{(dis)aligns} the speaker with the interlocutor. All three are expressed (often subtly) through overt linguistic forms, making stance observable at the surface level. The two turns above invert all three vectors: self-blame, hedges, filler, and an apologetic emoticon (AnxiousInquirer) versus company-blame, a negative evaluative epithet, and a bare imperative (FrustratedComplainer).}
\label{fig:stance-personas}
\end{figure}

Crucially, a social type is a \emph{relational} category, constituted in how a speaker positions themselves toward the talk and toward others, and not a demographic one, though it can intersect with identity categories without being reducible to them (e.g. \citet{podesva2007phonation}'s account of a queer speaker enacting a ``gay diva'' persona among close friends shows how the persona is assembled from stylistic choices such as falsetto phonation rather than derived from demographic membership). Elements of this view have begun to surface in simulation studies: for example, \citet{jeong2026locar} apply fine-grained sociolinguistic control to in-vehicle assistant simulation, and linguistic signals have been used to probe persona stability and collapse \citep{han2026personality, xiao2026chameleon}. Yet the persona is still typically modelled as something behind the words, rather than recognizing the words themselves as evidential (Figure~\ref{fig:stance-personas}). 

\subsection{Evaluating persona adherence}
\label{sec:related-adherence}
In task-oriented user simulation, the main fidelity signal largely remains an LLM-judge tasked with determining whether a simulated user is consistent with its persona, either through direct role-consistency scoring \citep{abdulhai2025consistently}, rubric-guided alignment \citep{liu2026muse}, persona-adaptive prompting of the evaluator \citep{purwar2026mmtp2}, or dedicated benchmarks for persona and role-play quality \citep{bao2026eval4sim, tu2024charactereval, peng2024quantifying}. Though LLM judges can be capable of nuanced text analysis, their best reported accuracy on a role-recognition task used to probe role-play alignment reaches only $\sim$69\%, compared to $\sim$91\% for human raters \citep{zhou2025personaeval}; they are also costly and require extensive calibration. As a step towards narrowing the "sim2real" gap \citep{zhou2026sim2real}, assessing user simulation faithfully requires instruments that are reproducible, deterministic, and able to localize where fidelity breaks down. However, deterministic, model-free measures rarely appear as the primary signal for persona fidelity. Aggregate human-likeness \citep{hathidara2026mirrorbench}, single-attribute knob-tracking \citep{chen2025recusersim}, or lexical analysis (e.g.\ LIWC, \citep{tausczik2010liwc}) have been offered as a descriptive lens to complement an LLM judge \citep{han2026personality}, though not as the headline metric. The stylometric techniques we adopt here have been developed in authorship verification and attribution \citep{aggazzotti2023speech, huang2024authorship, man2026explainable}, and in stylometric fingerprinting used to tell role-constrained LLM agents apart \citep{dietrich2026peers}; fidelity itself has been checked behaviourally against real user decisions \citep{chen2026simulated}, but not on the stylistic substrate where a persona is realized. To our knowledge, no prior work uses deterministic, spec-mapped stylometric measures as the primary fidelity signal for a multi-attribute persona in task-oriented customer-service simulation across text and voice. Because low fidelity can fall unevenly across user populations \citep{seshadri2026lost, haq2026dialect}, a reproducible way to detect and localize it is a prerequisite for trustworthy evaluation. While detecting a disparity is not the same as removing it, detection is a necessary first step toward any targeted intervention, especially for production systems that serve a linguistically diverse user base.

\section{Methodology}

\subsection{Evaluation system}
\label{sec:system}

Our study sits inside an end-to-end evaluation pipeline designed for task-oriented conversational tool-use agents. It consists of an LLM-based simulator impersonating the user, interacting with an agent harness. The agent harness has access to tool APIs, and a scoring layer with LLM-as-a-judge and deterministic checks aimed at its outputs. The user simulator is given a persona schema, a task scenario, and details specific to the scenario such as name or contact details. The simulator interacts with the agent via WebRTC and produces a multi-turn dialogue. Crucially, in this paper we hold the agent fixed and evaluate only the simulated user's turns in a transcript.

\subsection{Sociolinguistic personas}
\label{sec:sociolinguistic-personas}

The sociolinguistic persona is authored as a set of stance-linked stylistic features with corresponding instructions directing the LLM to instantiate them. We designed eight personas that instantiate example customer archetypes intended to be plausible across five service verticals (see ~\ref{sec:experiment}), with each persona's per-feature bands set by hand by a sociolinguist. An example persona schema is provided in Appendix~\ref{app:schemas}. In this schema (henceforth the \emph{definition} schema), a persona is specified as ordinal rate targets over eight linguistic features: question rate, message length, hedging, politeness markers, filler, all-caps, computer-mediated communication (CMC) acronyms such as \textit{lol, omg, wtf} etc., and affect valence. Each is then compiled to a band
($\textsc{none}<\textsc{low}<\textsc{medium}<\textsc{high}$) in the instruction prompt. We add per-modality targets for markers specific to text or voice (e.g.\ emoji and CMC acronym use are text-only). This schema also contains a ``behavioural\_contract'' block with a set of rules translating sociolinguistic markers into commands and rules counteracting the LLM's default assistant behaviour. \footnote{To assess whether fidelity gains can be solely attributed to stylistic rates, we ran an ablation test with this block removed; further discussion is included in Appendix~\ref{contracts}.} 
The \emph{archetype} baseline describes the same personas in the conventional way: as a set of defining adjectival traits, communication style, situational motivations, and example phrases. The A/B test isolates the measurable and instructive ``what to do'' versus the implicit and descriptive ``what you are'' persona design (\S\ref{sec:experiment}). Stating the target as an observable rate is what makes a persona both authorable at feature granularity and directly legible to the model-free metrics of \S\ref{sec:metrics}, where we also define how each observed rate is computed.

\subsection{Persona Fidelity metrics}
\label{sec:metrics}

\subsubsection{Stylometric distinguishability (authorship-verification ROC--AUC)}
\label{sec:metric-distinguishability}

We adapt \emph{authorship verification} as an unsupervised separability test, a framing that mirrors how \citet{bao2026eval4sim} operationalize simulator consistency. Each conversation's simulated-user turns are concatenated and fingerprinted with a character $n$-gram TF-IDF vector (\texttt{char\_wb}, $n\!\in\![2,4]$), a standard authorship-verification representation (as in the PAN shared tasks\footnote{https:\//pan.webis.de\/shared-tasks.html}) dominated by orthography, morphology, and function-word patterns rather than topics \citep{aggazzotti2023speech,huang2024authorship}. We compute the cosine similarity of every conversation pair, label each pair same- or different-persona, and report the ROC-AUC of similarity as a predictor of the same-persona label. AUC is the probability that a random same-persona pair is more similar than a random different-persona pair: $0.5$ denotes indistinguishable personas, and $1.0$ perfect separability. Because the statistic rewards both between-persona distinctiveness and within-persona consistency, a simulator that collapses to a bland neutral voice scores near chance. Crucially, the metric is feature-agnostic: it presupposes none of the authored style features (\S\ref{sec:metric-adherence}), providing an independent check that distinctiveness is not an artifact of the features we chose to author.

Four properties make the estimate more trustworthy. (i) A label-permutation control (shuffle persona ids, recompute AUC) collapses to $\approx\!0.5$, confirming the metric reads genuine persona identity rather than a shared baseline. (ii) Because every persona runs an identical scenario set, topic is crossed with persona identity and therefore cancels in the same-minus-different similarity gap the AUC rewards. (iii) The measurement is a fixed transform (character $n$-gram TF-IDF with cosine similarity) with no trained parameters, held identical across the \emph{archetype} and \emph{definition} arms, so any difference in AUC is attributable to the simulated text rather than the detector. (iv) We report a bootstrap $90\%$ CI over conversations and size the definition$-$archetype schema effect with a paired bootstrap on the AUC delta; alongside the AUC we report the within- versus between-persona similarity gap, which separates within-persona consistency from between-persona distinctiveness.

\subsubsection{Stylistic adherence (Kendall's $\tau_b$)}
\label{sec:metric-adherence}
Distinguishability certifies that personas differ, but it does not check that they differ as specified. Stylistic adherence is a rank-based test of spec-to-behaviour alignment. For each authored feature $f$, we map each persona's specified band ($\textsc{none}<\textsc{low}<\textsc{medium}<\textsc{high}$) to an ordinal, aggregate the observed rate of $f$ over that persona's
conversations, and compute Kendall's $\tau_b$ between the specified ordinal and the observed rate across personas. $\tau_b\!\in\![-1,1]$: $1$ means the observed rate climbs monotonically with the spec, $0$ no relationship, and negative an inversion (the persona does the opposite of its spec). We use $\tau_b$ for its tie-tolerance and robustness on the small per-persona $n$ typical of a pilot, and report Spearman's $\rho$ alongside. We compute a coarse per-persona variant (one point per persona) and a finer per-conversation variant (all conversations, reflecting within-persona spread), and summarize each arm by the mean $\tau_b$ over features with a varying specified level. Observed rates are extracted with rule-based counters for the structural features (question, message length, all-caps), small hand-authored micro-lexicons in the LIWC tradition for the closed-class features (hedging, politeness, CMC-acronym, filler) \citep{tausczik2010liwc}, and VADER for affect valence \citep{hutto2014vader}.

Together, the metrics answer complementary questions: are simulated personas \emph{distinct and self-consistent} (\S\ref{sec:metric-distinguishability}), and does the simulator \emph{realize each persona's specified style in the right direction} (\S\ref{sec:metric-adherence})? Because the metrics are sensitive to different things, reading them jointly allows us to localize where fidelity breaks down. And since they are deterministic measures, they are stable enough to hold as a fixed instrument while we vary what they measure.

\subsection{Experimental design}
\label{sec:experiment}

\paragraph{A/B factorial.} We use factorial A/B design to test our persona schema in a two-part experiment. First, we cross the persona \emph{schema} (archetype vs.\ definition) with the simulator \emph{model} in a $2\times2$, so a change in fidelity can be attributed to schema or model rather than confounded. Second, we ran the schema A/B in a \emph{voice} modality to test whether the effect replicates in a realtime voice pipeline. The schema factor's main effect measures how much of a fidelity gap is attributable to spec underspecification. The model factor's main effect tests whether a newer, more capable generation or higher reasoning effort makes it easier to steer, and the schema$\times$model interaction tests whether the richer schema pays off more for newer and stronger models. Finally, we include the voice mode to test whether the schema effect holds up under a different-modality pipeline.

\paragraph{Personas.} We test \nPersonas{} customer archetypes: the Anxious Inquirer, Busy Multitasker, Frustrated Complainer, Methodical Researcher, Social Chatterbox, Tech-Savvy Expert, Technophobe, and Vague Storyteller, shared across the two schemas under test. Each were chosen to span the stance and style dimensions encoded in the definition schema (\S\ref{sec:sociolinguistic-personas}). Across the set, affective baselines range from anxious and angry to calm and cheerful, epistemic stance from deferential and low-certainty to confident and authoritative, and each of the eight stylistic features from \textsc{none} to \textsc{high}.

\paragraph{Agents and corpus.} We run \nAgents{} task-oriented customer-service agents across 5 verticals (real estate, healthcare, agricultural, automotive, and technical-support), spanning tasks such as appointment booking, eligibility and identity checks, knowledge-base lookups, and request routing. Each bot interacted with \nPersonas{} distinct user personas. To ensure a consistent baseline, we drew from a larger gold-standard set to create a topic-matched, manually verified subset of 5 scenarios, which were reused across all personas and test arms, yielding \nPerPersona{} conversations per persona, per arm. We scrubbed the data of the conversations that failed due to technical errors such as connection or time out issues, and discarded arms with a high failure rate. This resulted in 4103 clean conversations across all arms.

\paragraph{Prompt encoding.} All 9 text models tested are instruction-tuned assistants and receive the identical role-play prompt: persona schema, scenario goal, scenario-specific information (e.g. desired appointment time), and user-profile content (e.g. name, email address, phone number, etc.; all synthetic), wrapped into the same explicit turn-taking and end-of-conversation directives. The voice arm takes the same input, but uses the \texttt{gpt-4o-realtime-preview} model's native speech pipeline with the default settings.

\section{Results}
\label{sec:results}

\subsection{Stylometric distinguishability}
\label{sec:results-distinguishability}
Distinguishability measures whether personas exhibit distinct stylistic variations from one another. The definition schema improves distinguishability AUC in text conversations for all \nAgents{} agents (Table~\ref{tab:results-text}), from a mean near the chance floor under the persona archetype baseline, to moderate distinguishability under the sociolinguistic schema, which hold in text and voice modalities.

\begin{table}[b]
  \centering
  \caption{Stylometric distinguishability (AUC) across text
  and voice, mean over \nAgents{} agents.}
  \label{tab:results-text}
  \resizebox{\columnwidth}{!}{%
  \begin{tabular}{lccc}
    \toprule
    & archetype & definition & $\Delta$ \\
    \midrule
    \multicolumn{4}{l}{\emph{text}} \\
    \quad \texttt{gpt-4o-mini}            & 0.483 & 0.593 & $+0.111$ \\
    \quad \texttt{gpt-5-mini}             & 0.449 & 0.636 & $+0.187$ \\
    \quad \texttt{gemini-3.5-flash (low)}       & 0.408 & 0.423 & $+0.015$ \\
    \quad \texttt{gemini-3.5-flash (high)}       & 0.481 & 0.482 & $+0.001$ \\
    \quad \texttt{gemini-3.1-pro-preview} & 0.417 & 0.414 & $-0.003$ \\
    \quad \texttt{gemma-4-31b-it}         & 0.403 & 0.426 & $+0.023$ \\
    \quad \texttt{claude-sonnet-5}        & 0.517 & 0.612 & $+0.095$ \\
    \quad \texttt{claude-haiku-4-5}       & 0.509 & 0.651 & $+0.141$ \\
    \quad \texttt{claude-opus-5}          & 0.565 & 0.670 & $+0.105$ \\
    
    \midrule
    \multicolumn{4}{l}{\emph{voice}} \\
    \quad \texttt{gpt-4o-realtime}  & 0.477 & 0.608 & $+0.131$ \\
    \bottomrule
  \end{tabular}}
\end{table}
Distinguishability scores by model show that stronger models are not necessarily more steerable, nor do they necessarily render the specified style at wider amplitude.\footnote{Notably, these results are not an artifact of model or persona verbosity: truncating every sample to a common character budget (down to 150 characters) leaves the overall definition schema advantage unchanged.} Though \texttt{claude-opus-5} shows the highest absolute distinguishability score for the definition arm out of all models tested, its archetype baseline was already the highest at the outset; it is also closely followed by the mid-tier \texttt{claude-haiku-4-5} and \texttt{gpt-5-mini}, and also bested by them in the size of the delta between the two schemas. When ranking Claude models by distinguishability gains, the lowest-ranked model by within-family capability --- \texttt{claude-haiku-4-5} --- shows the most improvement, followed then by \texttt{claude-opus-5}, and finally \texttt{claude-sonnet-5}. All three Gemini configurations perform poorly regardless of schema arm, and \texttt{gemini-3.1-pro-preview} is marginally \textit{worse} in the definition arm; all three also show a distinguishability score below chance for both arms. Higher reasoning effort in \texttt{gemini-3.5-flash} brings real gains in absolute scores but doesn't lift performance above chance. \texttt{gemma-4-31b-it} shows similar pattern, with the lowest archetype baseline of any model tested  and one of the smallest schema deltas. This suggests a model family or provider effect. Altogether, larger or more capable models are not necessarily more capable simulators than smaller, less advanced ones, highlighting the necessity of evaluating on multiple models from multiple families and classes. 

Inspection of the per-persona scores (see Appendix~\ref{app:perpersona}) also shows the gains are unevenly distributed. Personas defined primarily by \emph{presence} of specific stylistic features (methodical, technophobe, anxious) tend to separate the most, whereas those defined primarily by the absence of markers (\texttt{tech\_savvy\_expert}, \texttt{busy\_multitasker}) remain at the chance floor across models. Additionally, \texttt{social\_chatterbox} and \texttt{vague\_storyteller} \emph{converge} (both render as richly chatty and collapse toward one another, except for \texttt{claude-haiku-4-5}, which renders them more distinctly). The metrics thus earn their diagnostic value in highlighting sources of low separability: personas that share registers or lack distinctive fingerprints as a result of persona-definition or metric-sensitivity limits, versus differences in fidelity attributable to model capacity. 

\subsection{Stylistic adherence}
\label{sec:results-adherence}
Adherence measures whether a persona's behaviour aligns with its prescribed specification (e.g. a persona instructed to hedge most frequently does so). Under the definition schema, most models demonstrate strong adherence; the effect holds across both text and voice modalities. A notable exception is the Gemini family and Gemma, which stay essentially flat under both schemas. We confirm this in two ways. The per-persona evaluation averages the observed feature rate across all of a persona's conversations into a single aggregated score, testing whether the overall ordering matches the intended design. Per-conversation scoring treats each interaction as an independent observation tagged with its target specification level; this tests whether an individual conversation from a higher-specified persona consistently exhibits a higher feature rate than one from a lower-specified persona, without averaging.

\begin{table}[ht]
  \centering
  \caption{Stylistic adherence ($\tau_b$) across text and voice, mean over \nAgents{} agents. Voice is averaged over the six non-orthographic features it can realize; all-caps and CMC-acronyms are text-only (\S\ref{sec:metric-adherence}).}
  \label{tab:results-adherence}
  \resizebox{\columnwidth}{!}{%
  \begin{tabular}{lccc}
    \toprule
    & archetype & definition & $\Delta$ \\
    \midrule
    \multicolumn{4}{l}{\emph{per-persona}} \\
    \quad \texttt{gpt-4o-mini}            & 0.194 & 0.428 & $+0.235$ \\
    \quad \texttt{gpt-5-mini}             & 0.153 & 0.573 & $+0.420$ \\
    \quad \texttt{gemini-3.5-flash (low)}       & $-0.042$ & $-0.032$ & $+0.010$ \\
    \quad \texttt{gemini-3.5-flash (high)}       & $-0.041$ & 0.014 & $+0.054$ \\
    \quad \texttt{gemini-3.1-pro-preview} & 0.023 & 0.023 & $+0.000$ \\
    \quad \texttt{gemma-4-31b-it} & $-0.089$ & $-0.065$ & $+0.024$ \\
    \quad \texttt{claude-sonnet-5}        & 0.368 & 0.521 & $+0.153$ \\
    \quad \texttt{claude-haiku-4-5}       & 0.315 & 0.604 & $+0.289$ \\
    \quad \texttt{claude-opus-5}          & 0.236 & 0.634 & $+0.398$ \\
    \quad voice                           & 0.255 & 0.489 & $+0.234$ \\
    \midrule
    \multicolumn{4}{l}{\emph{per-conversation}} \\
    \quad \texttt{gpt-4o-mini}            & 0.104 & 0.304 & $+0.200$ \\
    \quad \texttt{gpt-5-mini}             & 0.092 & 0.485 & $+0.393$ \\
    \quad \texttt{gemini-3.5-flash (low)}       & $-0.020$ & $-0.030$ & $-0.010$ \\
    \quad \texttt{gemini-3.5-flash (high)}       & 0.004 & 0.007 & $+0.003$ \\
    \quad \texttt{gemini-3.1-pro-preview} & 0.012 & 0.005 & $-0.007$ \\
    \quad \texttt{gemma-4-31b-it} & $-0.029$ & $-0.020$ & $+0.008$ \\
    \quad \texttt{claude-sonnet-5}        & 0.232 & 0.405 & $+0.173$ \\
    \quad \texttt{claude-haiku-4-5}       & 0.191 & 0.499 & $+0.308$ \\
    \quad \texttt{claude-opus-5}          & 0.184 & 0.548 & $+0.364$ \\
    \quad voice                           & 0.139 & 0.380 & $+0.241$ \\
    \bottomrule
  \end{tabular}
  }
\end{table}

Adherence rises decisively both per-persona and per-conversation (the per-conversation numbers are lower only because a single conversation is noisier than an average) for all models except those in the Gemini family and Gemma. \texttt{gpt-5-mini} exhibits the highest gain and \texttt{claude-sonnet-5} the most modest. Voice --- averaged over the six non-orthographic features that it can realize, since all-caps and CMC-acronyms are text-only --- gains $+0.234$ per-persona, on par with the text models rather than an outlier; the apparent text advantage is carried by the two text-only markers (below), not by the modality. However, a per-feature breakdown shows that the gains are concentrated in fine-grained style markers (see Appendix~\ref{app:perfeature}), and the largest gains fall on the markers that a default assistant voice omits: CMC acronyms (absent under the archetype schema and restored under definition schema for OpenAI models,  up to $+0.838$ for \texttt{gpt-5-mini}) and all-caps (enacted \emph{opposite} to spec under the archetype for \texttt{gpt-4o-mini}, \texttt{gpt-5-mini}, and \texttt{claude-haiku-4-5}, which the schema flips positive, up to $+0.73$ for haiku). Features that a helpful register already exhibits, such as politeness, barely move. For voice, gains concentrate in the prosodic and disfluency markers the channel actually renders (hedging $+0.47$, filler $+0.42$). 

\section{Discussion and conclusion}
\label{sec:discussion}
\paragraph{Do the metrics transfer?} Both metrics yield actionable, internally consistent signal in simulated user persona fidelity measurement for task-oriented dialogue. They also detect what the design set out to test: the sociolinguistically-grounded definition schema (specifying each persona as concrete stylistic rates with `what to do` instructions) produced more faithful and stylistically distinct simulated users than a schema based on descriptive terms with `what you are` instructions, improving both spec adherence and inter-persona distinguishability over the baseline for six of \nModels models tested. They also enabled localization of the schema effect to specific stylistic features. A small scale human preference analysis (Appendix \ref{app:humanpref}) shows that the metrics generally align with human judgment at the extremes, with some exceptions among smaller effects.

\paragraph{Specifying persona sociolinguistically improves user simulator fidelity, but faithfulness $\neq$ naturalness.} Grounding personas in specific linguistic rates makes them inherently measurable and directly informs the prescriptive (``what to do'') instructions. Compared to broader descriptive archetypes (``who you are''), this combined schema produces more linguistically diverse and faithful renderings.  However, while the definition schema reliably raises scores on both metrics, qualitative analysis reveals a critical caveat: rate-based adherence certifies that a feature appears at the \emph{specified relative frequency and ordering}, not that it is realized \emph{naturally}. The model renders some markers formulaically, reducing politeness to repetitive ``Thank you so much!''\ / ``I hope that's okay!'', and disfluency to a predictable ``um'' at the start of a sentence. We can partially combat this by adding the measures of lexical diversity (e.g. Type-Token Ratio, Shannon Entropy) or conditioning them on specific dialog acts, but it remains difficult to capture the full nuance of persona authenticity with deterministic scoring only.

mulation; the metrics are a useful tool for determining which is most performant, and which makes a poor simulator.

\paragraph{The metrics work as a design instrument for authoring personas.} Part of the modest distinguishability also reflects genuine \emph{overlap} between personas. Qualitatively, a chatty ``social'' persona bleeds into a rambling ``storyteller,'' and an anxious persona into a technophobic one. They are distinct in specification but convergent in observable style. Thus, low pairwise separability flags persona definitions that should be pulled apart to maximize diversity, closing the loop from measurement back to authoring.

\paragraph{Conclusion.} Treating persona fidelity as a deterministic, spec-mapped diagnostic on observable style turns an opaque, judge-dependent quality signal into a feature-level, reproducible, and attributable one. The proposed metrics have proven most useful in error-attribution analysis, localizing where fidelity breaks down, as a first step towards interventions that would move user simulations towards more faithful renderings of diverse and variable linguistic outputs. That is both a prerequisite for trusting a simulator as a measurement instrument, and a concrete step toward narrowing the sim2real gap.

\section*{Limitations}
\label{sec:limitations}

\paragraph{Scale and scope:} This is a pilot study that covers \nAgents{} agents from a single domain (customer service) on one platform, with \nPersonas{} personas and \nPerPersona{} conversations per persona per arm; per-agent confidence intervals overlap, so the result rests on cross-agent consistency rather than any single significant effect.

\paragraph{Model coverage:} All models in our study are instruction-tuned and aligned to act as "helpful assistants". We can not determine whether this underlying alignment inherently limits achievable persona fidelity. We originally planned to test a user-aligned model \textsc{UserLM-8b} to isolate this effect, but it proved too unstable as a next-turn predictor and was left for future work.

\paragraph{Construct validity:} The metrics measure stylistic fidelity via eight engineered features and a character-$n$-gram fingerprint. This operationalization of ``style'' does not capture task/goal success, semantic appropriateness, or stylistic dimensions outside the feature set. Follow-up experiments will mine the sociolinguistic and stylometric literature for broader markers (e.g. intensification, discourse markers, syntactic complexity, etc.) and verify if our findings hold. We will also isolate the effects of specific stylistic markers and other parts of the spec, e.g. specific types of instructions.

\paragraph{Static persona specification:} The persona is supplied only once, as a prompt at the start of the conversation, with no in-dialogue reinforcement; the simulator must sustain it unaided across many turns, the very condition under which aligned models tend to drift back toward a default ``assistant'' voice \citep{xiao2026chameleon}. Our results therefore characterize fidelity under one-shot prompting. Testing live, in-conversation persona steering is reserved for future research.

\paragraph{Length-dependent distinguishability:} The ability to distinguish between personas relies heavily on conversation length. Mean transcript length correlates with the definition-arm AUC at r = 0.79 (and with the schema effect at r = 0.73) across the five agents. Longer conversations naturally provide more opportunities to drop stylistic markers, and thus  distinguishability is driven both by the persona schema and the speech the agent elicits from the persona. Follow-up experiments will incorporate length normalization to better isolate this effect.

\section*{Ethics Statement}
The paper discusses simulation fidelity using LLM-simulated users and hand-authored lexicons, with no human subjects data involved. We access the assets used in this study according to their stated licenses and providers' APIs. Annotation in the preference alignment study was performed in-house, without external annotation services.
\bibliographystyle{acl_natbib}
\bibliography{references}

\appendix

\section{Example persona schemas: archetype vs.\ definition}
\label{app:schemas}
To illustrate the contrast between the two authored schemas, we show both for a single
persona, \emph{The Anxious Inquirer} (referenced in \S\ref{sec:related}). The
\textbf{archetype} schema specifies the persona as flat adjective labels, a prose
description, and example phrases; the \textbf{definition} schema specifies it as a stance profile plus modality-specific ordinal style \emph{rates}, which are then compiled into bands. Both are the verbatim inputs supplied to the simulator in A/B simulations.

\subsection{Archetype schema (label-based baseline)}
\begin{lstlisting}
{
  "archetype_name": "The_Anxious_Inquirer",
  "description": "This user is hesitant and nervous, seeking constant
    confirmation and reassurance before proceeding with any action.",
  "defining_traits": ["Hesitant", "Repetitive", "Seeks reassurance",
    "Apologetic"],
  "communication_style": {
    "conciseness": "Medium", "formality": "High",
    "use_of_jargon": "None", "question_frequency": "High"
  },
  "situational_motivations": [
    "Avoid making a costly mistake",
    "Ensure they fully understand every step",
    "Gain confidence in their decision"
  ],
  "example_phrases": [
    "Are you sure this is the right button to click?",
    "Can you confirm that my data is safe if I proceed?",
    "So if I do this, nothing bad will happen, right?",
    "I'm sorry to ask so many questions, but I just want to be certain."
  ]
}
\end{lstlisting}

\subsection{Definition schema (sociolinguistic, rate-based)}
\begin{lstlisting}
{
  "persona_id": "anxious_inquirer",
  "source_archetype": "The_Anxious_Inquirer",
  "identity": {
    "backstory": "A cautious, anxious person who worries about getting
      things wrong and needs reassurance before acting; not confident
      with technology; tends to over-explain and double-check."
  },
  "stance_profile": {
    "affective": {
      "baseline_state": "anxious", "intensity": 6,
      "frustration_baseline": 3,
      "triggers": {"vague_answer": 4, "rushing": 3,
        "repeated_question": 4, "policy_barrier": 5}
    },
    "epistemic": {"certainty": 2, "authority_claims": "none"},
    "investment": {
      "stake": 8, "persistence": 8,
      "withholding_strategy": ["shares full details only when reassured"]
    },
    "alignment": {
      "convergence_tendency": 7,
      "asymmetry": {"gain_per_positive": 1.0, "loss_per_negative": 1.5}
    }
  },
  "linguistic_style": {
    "shared": {
      "register_tenor": "apologetic", "jargon_rate": "none",
      "hedging_rate": "high", "politeness_marker_rate": "high",
      "question_frequency": "high", "verbosity": "medium"
    },
    "text": {
      "emoji_rate": "none", "emoticon_rate": "high",
      "all_caps_rate": "none", "cmc_acronym_rate": "low",
      "ellipsis_rate": "high", "filler_rate": "high"
    },
    "voice": {
      "pitch": "high", "pace": "fast", "uptalk": "high",
      "interruption_tendency": 0.1, "backchannel_tendency": 0.5,
      "disfluency_tendency": 0.8, "yield_resistance": 0.1,
      "backchannel_markers": ["uh-huh", "mm-hmm", "mm"]
    }
  },
  "behavioral_contracts": {
    "MUST": [
      "ask for confirmation before accepting each step",
      "express relief when reassured",
      "over-explain in long, run-on sentences, sometimes dropping words"
    ],
    "MUST_NOT": [
      "assert confident knowledge of policy",
      "accept a resolution silently without checking"
    ]
  }
}
\end{lstlisting}

\section{Ablation test without behavioural contracts block}
\label{contracts}
To assess their impact on fidelity gains, we performed a simple ablation test removing the \texttt{behavioural\_contracts} (MUST / MUST\_NOT) block from the definition schema while holding everything else fixed. We performed this exploratory test only on the most performant model in the set, \texttt{gpt-5-mini}, against all \nAgents{} agents, scored on the subset of conversations clean in both variants. Pairwise comparisons between the fidelity metric scores for the ablated conversation set and the original \texttt{gpt-5-mini} set with the behavioural contracts block are summarized in Table ~\ref{tab:ablation}; the baseline archetype schema scores are also included for reference. The delta score represents the difference between definition runs, i.e. the no block score --- the with block score. 

Removing the block lowers every metric score relative to the conversations with the block; it is clear that for the assistant-aligned models tested in this study, the behavioural contracts block does contribute to the realizations of stylistic features.  However, the no block scores are still significantly higher than the archetype baseline (p = 0.031, Wilcoxon p = 0.031, one-sided), indicating that even the rate specifications alone already lift the metric scores substantially. The rates are also what make style observable and quantifiable --- and, ultimately, extractable from the source transcripts. Regardless, neutralizing the assistant behaviour explicitly is still necessary.  The drop in scores suggests that the behavioural contracts block is doing de-alignment work that prevents the model from stylistic regression; without it, the agent will still default to its helpful-assistant voice and render each persona in a similar style. The feature dials do still move the rates, but sociolinguistic behavioural instruction is still required. In future experiments, we will test whether a model specifically aligned for user roleplays will likewise require this additional instruction.

\begin{table}[ht]
  \centering
  \caption{Ablation of the behavioural-contracts block, \texttt{gpt-5-mini}.}
  \label{tab:ablation}
  \resizebox{\columnwidth}{!}{
  \begin{tabular}{lcccc}
    \toprule
    metric & archetype & no block & with block & $\Delta$ \\
    \midrule
    Distinguishability AUC                & 0.453 & 0.518 & 0.652 & $-0.134$ \\
    Adherence $\tau_b$ (per-persona)      & 0.181 & 0.489 & 0.619 & $-0.130$ \\
    Adherence $\tau_b$ (per-conversation) & 0.106 & 0.379 & 0.515 & $-0.136$ \\
    \bottomrule
  \end{tabular}}
\end{table}

\section{Per-persona distinguishability}
\label{app:perpersona} 
Table~\ref{tab:app-perpersona} decomposes the corpus distinguishability AUC
(\S\ref{sec:results-distinguishability}) into its per-persona sources. For each persona we report three quantities: \emph{same}, the mean similarity among that persona's own conversations (within-persona consistency); \emph{diff}, the mean similarity between that persona's conversations and those of every other persona (between-persona overlap); and \emph{gap}, the per-persona separability determined by~\emph{same}~$-$~\emph{diff}. A large positive gap marks a sharply drawn persona whose conversations cluster together; a gap at or below zero marks one the metric cannot separate from the others. Because all conversations share a customer-service register, absolute similarities are uniformly high, so the persona-specific signal is carried by the gap value. Tables~\ref{tab:app-perpersona-new} and~\ref{tab:app-perpersona-new2} report the same per-persona decomposition for the Gemini and Claude models.

\begin{table*}[tp]
  \centering
  \caption{Per-persona separability (\textsc{definition} arm), OpenAI models, mean over \nAgents{} agents.}
  \label{tab:app-perpersona}
  \footnotesize
  \begin{tabular}{lccccccccc}
    \toprule
    & \multicolumn{3}{c}{\texttt{gpt-4o-mini}} & \multicolumn{3}{c}{\texttt{gpt-5-mini}} & \multicolumn{3}{c}{voice} \\
    \cmidrule(lr){2-4}\cmidrule(lr){5-7}\cmidrule(lr){8-10}
    Persona & same & diff & gap & same & diff & gap & same & diff & gap \\
    \midrule
    anxious\_inquirer      & 0.618 & 0.469 & $+0.149$ & 0.648 & 0.482 & $+0.166$ & 0.640 & 0.531 & $+0.109$ \\
    vague\_storyteller     & 0.592 & 0.460 & $+0.131$ & 0.614 & 0.478 & $+0.135$ & 0.677 & 0.529 & $+0.148$ \\
    methodical\_researcher & 0.512 & 0.439 & $+0.073$ & 0.582 & 0.462 & $+0.120$ & 0.535 & 0.484 & $+0.051$ \\
    technophobe            & 0.465 & 0.439 & $+0.026$ & 0.528 & 0.442 & $+0.087$ & 0.575 & 0.501 & $+0.074$ \\
    social\_chatterbox     & 0.545 & 0.449 & $+0.096$ & 0.538 & 0.458 & $+0.080$ & 0.610 & 0.515 & $+0.096$ \\
    frustrated\_complainer & 0.406 & 0.393 & $+0.013$ & 0.448 & 0.423 & $+0.026$ & 0.475 & 0.454 & $+0.021$ \\
    busy\_multitasker      & 0.234 & 0.291 & $-0.057$ & 0.269 & 0.290 & $-0.021$ & 0.274 & 0.328 & $-0.054$ \\
    tech\_savvy\_expert    & 0.311 & 0.362 & $-0.051$ & 0.334 & 0.380 & $-0.047$ & 0.375 & 0.407 & $-0.033$ \\
    \bottomrule
  \end{tabular}
\end{table*}
\begin{table*}[tp]
  \centering
  \caption{Per-persona separability (\textsc{definition} arm), Gemini and Gemma models, mean over \nAgents{} agents.}
  \label{tab:app-perpersona-new}
  \resizebox{\textwidth}{!}{%
  \footnotesize
  \begin{tabular}{lcccccccccccc}
    \toprule
    & \multicolumn{3}{c}{\texttt{gemini-3.5-flash (low)}} & \multicolumn{3}{c}{\texttt{gemini-3.1-pro-preview}} &
    \multicolumn{3}{c}{\texttt{gemini-3.5-flash (high)}} & \multicolumn{3}{c}{\texttt{gemma-4-31b-it }}\\
    \cmidrule(lr){2-4}\cmidrule(lr){5-7}\cmidrule(lr){8-10}\cmidrule(lr){11-13}
    Persona & same & diff & gap & same & diff & gap & same & diff & gap & same & diff & gap\\
    \midrule
    anxious\_inquirer      & 0.323 & 0.357 & $-0.034$ & 0.327 & 0.372 & $-0.045$ & 0.238 & 0.250 & $-0.012$ & 0.189 & 0.200 & $-0.012$\\
    vague\_storyteller     & 0.303 & 0.356 & $-0.053$ & 0.347 & 0.382 & $-0.035$ & 0.221 & 0.244 & $-0.022$ & 0.173 & 0.194 & $-0.020$ \\
    methodical\_researcher & 0.318 & 0.360 & $-0.042$ & 0.327 & 0.373 & $-0.045$ & 0.229 & 0.247 & $-0.018$ & 0.178 & 0.197 & $-0.019$\\
    technophobe            & 0.347 & 0.376 & $-0.028$ & 0.332 & 0.377 & $-0.045$ & 0.243 & 0.254 & $-0.011$ & 0.195 & 0.203 & $-0.008$\\
    social\_chatterbox     & 0.307 & 0.361 & $-0.054$ & 0.308 & 0.370 & $-0.062$ & 0.246 & 0.254 & $-0.008$ & 0.169 & 0.193 & $-0.023$ \\
    frustrated\_complainer & 0.321 & 0.364 & $-0.043$ & 0.335 & 0.382 & $-0.047$ & 0.242 & 0.253 & $-0.011$ & 0.187 & 0.200 & $-0.013$\\
    busy\_multitasker      & 0.323 & 0.364 & $-0.041$ & 0.329 & 0.376 & $-0.047$ & 0.249 & 0.255 & $-0.006$ & 0.181 & 0.198 & $-0.017$\\
    tech\_savvy\_expert    & 0.301 & 0.357 & $-0.056$ & 0.339 & 0.382 & $-0.042$ & 0.241 & 0.253 & $-0.012$ & 0.190 & 0.200 & $-0.010$\\
    \bottomrule
  \end{tabular}%
  }
\end{table*}

\begin{table*}[tp]
  \centering
  \caption{Per-persona separability (\textsc{definition} arm), Claude models, mean over \nAgents{} agents.}
  \label{tab:app-perpersona-new2}
  \footnotesize
  \begin{tabular}{lccccccccc}
    \toprule
    & \multicolumn{3}{c}{\texttt{claude-sonnet-5}} & \multicolumn{3}{c}{\texttt{claude-haiku-4-5}} & \multicolumn{3}{c}{\texttt{claude-opus-5}} \\
    \cmidrule(lr){2-4}\cmidrule(lr){5-7}\cmidrule(lr){8-10}
    Persona & same & diff & gap & same & diff & gap & same & diff & gap \\
    \midrule
    anxious\_inquirer      & 0.540 & 0.387 & $+0.153$ & 0.637 & 0.490 & $+0.147$ & 0.595 & 0.381 & $+0.214$ \\
    vague\_storyteller     & 0.438 & 0.379 & $+0.059$ & 0.663 & 0.502 & $+0.161$ & 0.469 & 0.363 & $+0.106$ \\
    methodical\_researcher & 0.401 & 0.363 & $+0.037$ & 0.516 & 0.448 & $+0.068$ & 0.476 & 0.364 & $+0.112$ \\
    technophobe            & 0.506 & 0.378 & $+0.128$ & 0.597 & 0.489 & $+0.107$ & 0.543 & 0.379 & $+0.165$ \\
    social\_chatterbox     & 0.416 & 0.362 & $+0.054$ & 0.661 & 0.496 & $+0.165$ & 0.475 & 0.342 & $+0.134$ \\
    frustrated\_complainer & 0.355 & 0.338 & $+0.017$ & 0.494 & 0.440 & $+0.054$ & 0.334 & 0.307 & $+0.027$ \\
    busy\_multitasker      & 0.241 & 0.282 & $-0.041$ & 0.267 & 0.316 & $-0.049$ & 0.193 & 0.238 & $-0.045$ \\
    tech\_savvy\_expert    & 0.251 & 0.287 & $-0.037$ & 0.312 & 0.351 & $-0.039$ & 0.234 & 0.256 & $-0.022$ \\
    \bottomrule
  \end{tabular}
\end{table*}

\FloatBarrier
\section{Per-feature adherence}
\label{app:perfeature} 
Tables~\ref{tab:results-features}, \ref{tab:results-features-new}, and~\ref{tab:results-features-new2} give the per-persona Kendall's $\tau_b$  (\S\ref{sec:results-adherence}) for each of the eight authored features (archetype vs.\ definition and their gap) across simulator models. Note that CMC acronyms and all-caps are text-only orthographic markers, and are reported as N/A for the voice channel.

\begin{table*}[tp]
  \centering
  \caption{Per-feature adherence ($\tau_b$, per-persona), OpenAI models, mean over \nAgents{} agents.}
  \label{tab:results-features}
  \footnotesize
  \begin{tabular}{lccccccccc}
    \toprule
    & \multicolumn{3}{c}{\texttt{gpt-4o-mini}} & \multicolumn{3}{c}{\texttt{gpt-5-mini}} & \multicolumn{3}{c}{voice} \\
    \cmidrule(lr){2-4}\cmidrule(lr){5-7}\cmidrule(lr){8-10}
    Feature & arch. & def. & $\Delta$ & arch. & def. & $\Delta$ & arch. & def. & $\Delta$ \\
    \midrule
    CMC-acronym rate & $0.000$  & $0.305$ & $+0.305$ & $0.000$  & $0.838$ & $+0.838$ & N/A & N/A & N/A \\
    filler rate      & $0.121$  & $0.654$ & $+0.533$ & $0.033$  & $0.653$ & $+0.620$ & $0.148$ & $0.567$  & $+0.418$ \\
    all-caps rate    & $-0.238$ & $0.154$ & $+0.393$ & $-0.126$ & $0.493$ & $+0.619$ & N/A & N/A & N/A \\
    hedging rate     & $0.203$  & $0.518$ & $+0.315$ & $0.071$  & $0.680$ & $+0.608$ & $0.119$ & $0.587$  & $+0.468$ \\
    message length   & $0.652$  & $0.668$ & $+0.016$ & $0.338$  & $0.734$ & $+0.396$ & $0.619$ & $0.751$  & $+0.132$ \\
    politeness rate  & $0.338$  & $0.619$ & $+0.280$ & $0.602$  & $0.737$ & $+0.135$ & $0.470$ & $0.619$  & $+0.148$ \\
    question rate    & $0.165$  & $0.215$ & $+0.050$ & $0.078$  & $0.191$ & $+0.113$ & $0.026$ & $0.182$  & $+0.156$ \\
    affect valence   & $0.307$  & $0.291$ & $-0.016$ & $0.229$  & $0.260$ & $+0.032$ & $0.150$ & $0.230$  & $+0.081$ \\
    \bottomrule
  \end{tabular}
\end{table*}
\begin{table*}[tp]
  \centering
  \caption{Per-feature adherence ($\tau_b$, per-persona), Gemini and Gemma models, mean over \nAgents{} agents.}
  \label{tab:results-features-new}
  \resizebox{\textwidth}{!}{%
  \footnotesize
  \begin{tabular}{lcccccccccccc}
    \toprule
    & \multicolumn{3}{c}{\texttt{gemini-3.5-flash (low)}} & \multicolumn{3}{c}{\texttt{gemini-3.1-pro-preview}} & \multicolumn{3}{c}{\texttt{gemini-3.5-flash (high)}} & \multicolumn{3}{c}{\texttt{gemma-4-31b-it}}\\
    \cmidrule(lr){2-4}\cmidrule(lr){5-7}\cmidrule(lr){8-10}\cmidrule(lr){11-13}
    Feature & arch. & def. & $\Delta$ & arch. & def. & $\Delta$ & arch. & def. & $\Delta$ & arch. & def. & $\Delta$\\
    \midrule
    CMC-acronym rate & $0.000$  & $0.000$  & $0.000$  & $0.000$  & $0.000$  & $0.000$ & $0.000$  & $0.000$  & $0.000$ & $0.000$  & $0.000$  & $0.000$ \\
    filler rate      & $0.000$  & $0.000$  & $0.000$  & $0.000$  & $0.000$  & $0.000$ & $0.000$  & $0.000$  & $0.000$ & $0.000$  & $0.000$  & $0.000$ \\
    all-caps rate    & 0.054  & $-0.199$ & $-0.253$ & $0.199$  & $-0.011$ & $-0.210$ & 0.577 & 0.157 & $-0.420$ & $-0.157$ & $-0.262$ & $-0.105$ \\
    hedging rate     & $0.080$  & $-0.035$ & $-0.115$ & $0.154$  & $-0.120$ & $-0.274$ & $-0.276$ & $0.276$ & $+0.552$ & $-0.197$ & 0.267 & $+0.464$ \\
    message length   & $-0.173$ & $0.008$  & $+0.182$ & $-0.142$ & $0.299$  & $+0.441$ & $-0.619$ & $-0.041$ & $+0.578$ & $-0.371$ & $-0.041$ & $+0.330$ \\
    politeness rate  & $0.074$  & $-0.008$ & $-0.083$ & $-0.124$ & $0.107$  & $+0.231$ & 0.124 & 0.124 & 0.000 & 0.041 & 0.371 & $+0.330$ \\
    question rate    & 0.001  & $-0.106$ & $-0.107$ & $0.027$  & $-0.091$ & $-0.118$ & $-0.564$ & $-0.130$ & $+0.434$ &  -0.043 & $-0.217$ & $-0.174$ \\
    affect valence   & $-0.370$ & 0.087  & $+0.457$ & $0.067$  & $0.000$  & $-0.067$ & 0.433 & $-0.276$ & $-0.709$ & $-0.670$ & $-0.749$ & $-0.079$ \\
    \bottomrule
  \end{tabular}
  }
\end{table*}

\begin{table*}[tp]
  \centering
  \caption{Per-feature adherence ($\tau_b$, per-persona), Claude models, mean over \nAgents{} agents.}
  \label{tab:results-features-new2}
  \footnotesize
  \begin{tabular}{lccccccccc}
    \toprule
    & \multicolumn{3}{c}{\texttt{claude-sonnet-5}} & \multicolumn{3}{c}{\texttt{claude-haiku-4-5}} & \multicolumn{3}{c}{\texttt{claude-opus-5}} \\
    \cmidrule(lr){2-4}\cmidrule(lr){5-7}\cmidrule(lr){8-10}
    Feature & arch. & def. & $\Delta$ & arch. & def. & $\Delta$ & arch. & def. & $\Delta$ \\
    \midrule
    CMC-acronym rate & $0.189$ & $0.564$ & $+0.375$ & $0.284$  & $0.631$ & $+0.347$ & $-0.032$ & $0.556$ & $+0.588$ \\
    filler rate      & $0.593$ & $0.577$ & $-0.015$ & $0.580$  & $0.723$ & $+0.144$ & $0.087$  & $0.548$ & $+0.461$ \\
    all-caps rate    & $0.072$ & $0.367$ & $+0.295$ & $-0.156$ & $0.573$ & $+0.729$ & $0.106$  & $0.533$ & $+0.427$ \\
    hedging rate     & $0.235$ & $0.652$ & $+0.418$ & $0.515$  & $0.782$ & $+0.267$ & $0.080$  & $0.767$ & $+0.687$ \\
    message length   & $0.668$ & $0.635$ & $-0.033$ & $0.371$  & $0.619$ & $+0.247$ & $0.437$  & $0.833$ & $+0.396$ \\
    politeness rate  & $0.437$ & $0.638$ & $+0.200$ & $0.437$  & $0.696$ & $+0.259$ & $0.569$  & $0.726$ & $+0.157$ \\
    question rate    & $0.398$ & $0.442$ & $+0.044$ & $0.261$  & $0.425$ & $+0.164$ & $0.243$  & $0.442$ & $+0.199$ \\
    affect valence   & $0.354$ & $0.292$ & $-0.063$ & $0.229$  & $0.386$ & $+0.157$ & $0.400$  & $0.667$ & $+0.267$ \\
    \bottomrule
  \end{tabular}
\end{table*}

\FloatBarrier

\section{Human Preference Analysis}
\label{app:humanpref}
We run a small scale blinded human preference analysis (archetype vs. definition vs. no preference; 10 randomly sampled transcript pairs per model), with two annotators rating pairs of transcripts produced by the two persona schemas under otherwise identical settings.

\paragraph{Per-model trends.} Directionally, human preference tracked the metrics at the extremes. For the two strongest effects on the automated metrics -- \texttt{claude-opus-5} ($\Delta_{\text{AUC}}=+0.105$, the largest adherence gain in the study) and \texttt{claude-sonnet-5} ($\Delta_{\text{AUC}}=+0.095$) -- annotators expressed a definite preference in most cases, favoring the definition schema. At the other end, \texttt{gemini-3.5-flash (low)}'s near-zero effect corresponds to "no preference" in 85\% of cases. An interesting exception is \texttt{gemini-3.5-flash (high)}, where the automated AUC delta was flat ($+0.001$) but human annotators expressed a definite preference in 62\% of cases and favored the definition arm.
\paragraph{Per-persona trends.}
The two personas with lowest distinguishability (\texttt{tech\_savvy\_expert} and \texttt{busy\_multitasker}) across every model were also rated as mostly "no preference" by the annotators. Interestingly, \texttt{frustrated\_complainer}, with the lowest "no preference" rate, leaned towards archetype schema while the automated metrics showed modest gains for definition schema. The sample size for each persona is small and inconsistent, and these findings should be taken with a grain of salt.

\end{document}